\pdfoutput=1

\documentclass[11pt]{article}

\usepackage[preprint]{acl}

\usepackage{times}
\usepackage{latexsym}

\usepackage[T1]{fontenc}

\usepackage[utf8]{inputenc}
\usepackage{paralist}

\usepackage{microtype}

\usepackage{inconsolata}
\usepackage{graphicx}
\usepackage{subfig}
\usepackage{multirow}
\usepackage{amsmath}
\usepackage{amsfonts} 
\usepackage{bbding}
\usepackage{makecell}
\usepackage{colortbl}
\usepackage{xcolor}
\usepackage{microtype}
\usepackage{bm}
\usepackage{booktabs}
\usepackage{wrapfig}
\usepackage{subfiles}
\usepackage{tcolorbox}
\usepackage{tikz}
\definecolor{cred}{HTML}{FF6B6B}
\definecolor{cyellow}{HTML}{FEC260}
\definecolor{cgreen}{HTML}{70AD47}
\definecolor{cblue}{HTML}{4D96FF}
\definecolor{cpurple}{HTML}{2A0944}
\definecolor{ggray}{RGB}{127,127,127}
\definecolor{aliceblue}{rgb}{0.94, 0.97, 1.0}

\newcommand{\eg}{\textit{e}.\textit{g}., }
\usepackage{color}

\usepackage{colortbl}  
\usepackage{xcolor}
\usepackage{pifont}
\usepackage{amsfonts}
\usepackage{soul}
\usepackage{enumitem}
\usepackage{bm}
\makeatletter
\newcommand{\ssymbol}[1]{$^{\@fnsymbol{#1}}$}
\makeatother

\newcommand{\tabfootnotesize}{\fontsize{8}{9}\selectfont}

\newcommand{\myparagraph}[1]{\textbf{#1}\hspace{1.8ex}}

\definecolor{myblue}{RGB}{218,232,252}
\definecolor{mygray}{RGB}{220,220,220}
\definecolor{mypink}{RGB}{251,49,153}

\newcommand{\VarSty}[1]{\textnormal{\ttfamily\color{blue!90!black}#1}\unskip}

\usepackage{array} 
\newcolumntype{x}[1]{>{\centering\arraybackslash}p{#1pt}}

\newlength\savewidth

\title{
\textit{Evolver}: Chain-of-Evolution Prompting to Boost Large Multimodal Models for Hateful Meme Detection}
\author{Jinfa Huang\textsuperscript{1}$^{*}$,  Jinsheng Pan\textsuperscript{1}\thanks{Equal contribution.}, Zhongwei Wan\textsuperscript{2}, Hanjia Lyu\textsuperscript{1}, Jiebo Luo\textsuperscript{1} \\
\textsuperscript{1}University of Rochester \quad
\textsuperscript{2}The Ohio State University \\
\{jhuang90, jpan24, hlyu5\}@ur.rochester.edu, \\
wan.512@osu.edu, \quad jluo@cs.rochester.edu \\
\textbf{\url{https://github.com/inFaaa/Evolver}} \\
}

\begin{document}
\maketitle
\begin{abstract}
Hateful memes continuously evolve as new ones emerge by blending progressive cultural ideas, rendering existing methods that rely on extensive training obsolete or ineffective. 
In this work, we propose \textit{Evolver}, which incorporates Large Multimodal Models (LMMs) via Chain-of-Evolution (CoE) Prompting, by integrating the evolution attribute and in-context information of memes. Specifically, \textit{Evolver} simulates the evolving and expressing process of memes and reasons through LMMs in a step-by-step manner using an evolutionary pair mining module, an evolutionary information extractor, and a contextual relevance amplifier. 
Extensive experiments on public FHM, MAMI, and HarM datasets show that CoE prompting can be incorporated into existing LMMs to improve their performance. More encouragingly, it can serve as an interpretive tool to promote the understanding of meme evolution.

\textcolor{red}{\textbf{Disclaimer.} This paper contains offensive content that may be disturbing to some readers.}
\end{abstract}

\section{Introduction}

\begin{figure}[ht]
\centering
    \includegraphics[width=1.0\linewidth]{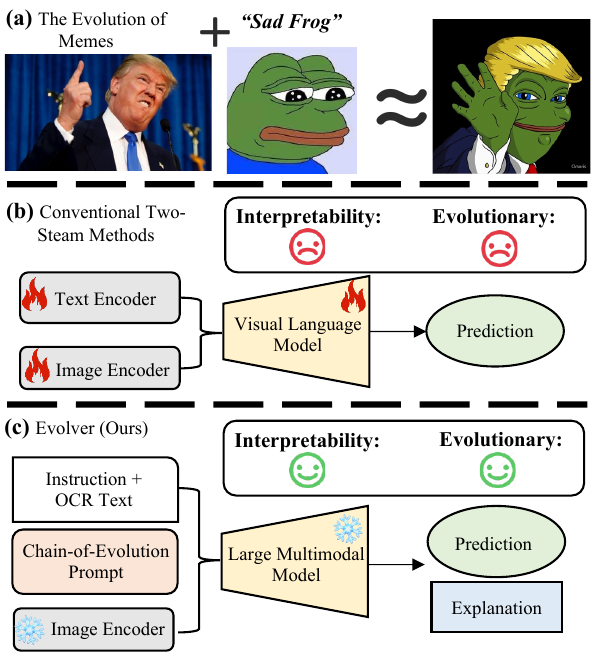}     
\vskip -0.1in
\caption{\textbf{ The illustration of (a) the evolution of memes and comparison between (b) conventional two-stream methods, and (c) our \textit{Evolver} method.} Memes evolve by fusing new cultural concepts. The meme of Trump is influenced by the meme of a sad frog in an image and text symbol, which creates a new hateful meme. Conventional hateful meme detection methods use trainable two-stream encoders and fusion for meme classification, with poor interpretability.  In contrast, our \textit{Evolver} captures the evolution and context of memes, utilizing them as prompts for large multimodal models to obtain a comprehensive understanding of memes.
}
\label{fig:intro}
\vskip -0.15in
\end{figure}

Hateful meme detection~\cite{lippe2020multimodal,kiela2020hateful,cao2020deephate} is a crucial task in the field of multimodal research, aiming to identify content that combines text and images to propagate hate speech or offensive messages. 
Memes, as a widespread cultural phenomenon, proliferate on the Internet, blending images and texts to convey sophisticated meanings. 
Furthermore, the fusion of visual and textual elements in memes complicates the interpretation of their semantics and the identification of hateful undertones.
The combination of text and images presents significant challenges in hateful meme detection, especially for detecting and moderating hateful content~\cite{levine2013controversial}. 

With the advances in image-text pre-training, efforts to leverage and fine-tune CLIP~\cite{radford2021learning} for hateful meme detection have demonstrated notable success in achieving high accuracy~\cite{kumar2022hate,arya2024multimodal,hee2022explaining}. 
However, as shown in Figure~\ref{fig:intro}(a), the extensive evolution of memes fusing together  
complicates the detection of hateful memes as they continuously evolve, bring new cultural elements to form new expressions.
Traditional methods for hateful meme detection, as illustrated in Figure~\ref{fig:intro}(b), suffer from limitations in interpretability and adaptability. 
These shortcomings hinder the understanding of memes' evolving nature and their contextual nuances, leading to overfitting on training sets and diminished effectiveness in the dynamic and evolving  meme landscape.

To overcome this dilemma and inspired by ~\cite{dawkins2016selfish,dawkins2016extended} which shows that similar biological groups share some common traits, we introduce \textit{Evolver}, a novel approach that uses Large Multimodal Models (LMMs)~\cite{li2023multimodal,lyu2023gpt,yang2023dawn} for hateful meme detection. By developing a benchmark specifically tailored for hateful meme detection based on LMMs, we aim to address the aforementioned limitations of traditional detection methods. Our approach not only enhances detection capabilities but also provides a more explainable and adaptable framework suitable for the evolving landscape of Internet memes. \textbf{Evolver} is an innovative approach in zero-shot hateful meme detection significantly. Extensive experiments have shown the effectiveness of our method across three widely recognized hateful meme detection datasets, demonstrating a superior ability to identify and interpret hateful memes.

Overall, the main contributions of this work are:
\begin{itemize}[topsep=0pt, partopsep=0pt, leftmargin=13pt, parsep=0pt, itemsep=3pt]
\item We establish an LMM-based zero-shot hateful meme detection benchmark, which provides a comprehensive evaluation of the application of LMMs in social media.
\item We propose a simple yet effective \textit{Evolver} framework, which advances LMMs with Chain-of-Evolution prompting. It expands LMMs with an evolution reasoning ability while offering good interpretability.
\item Extensive experiments on commonly used zero-shot hateful meme detection datasets with superior performance validate the efficacy and generalization of our method.
\end{itemize}
\section{Approach}


\subsection{Problem Definition} We define the task of hateful meme detection as a binary classification task. Meme, in our case, consists of an image-text pair. This task can be illustrated as follows: 
\begin{equation}
    \hat y_j = g(\{X^{j}_v, X^{j}_t\})     
\end{equation} where $\hat y_j$ denotes the \textit{j}-th prediction and the prediction $\hat y_j \in \{0, 1\}$ indicates the target image-text pair is hateful or not. $g(\cdot)$ is the multimodal model. $\{X^{j}_v, X^{j}_t\}$ is the \textit{j}-th image-text paired input.  


\subsection{Large Multimodal Models}


\myparagraph{Vision Encoder}
serves as a translator to help language models understand visual content. It leverages frozen pretrained vision models such as CLIP~\cite{radford2021learning} and ViT~\cite{dosovitskiy2020image} to encode visual content so that the language model can understand visual content: 
\begin{equation}
        h_v = \bm{W} \cdot {\text{Enc}_{vis}} (X_v)
\end{equation}
where $h_v$ is the language embedding tokens. $\bm{W}$ is the projection to transform visual features into language embedding tokens. $\text{Enc}_{vis}(X_v)$ denotes the visual feature extracted by pretrained model. 

\noindent \myparagraph{Large Language Model Decoder} generates a sentence given tokens~\cite{touvron2023llama}. The process of generation can be represented as: 
\begin{gather}
    h_t = \text{Tokenize} (X_t) \\ 
    p(w) = \prod^n_{i=1} p(w_i | w_{<i}, h_t, h_v)
\end{gather}where $X_t$ is the input text, $\text{Tokenize}(\cdot)$ transform text into tokens, $h_v$ is the image tokens, $p(w)$ is the probability of generating a sentence by a language model, $p(w_i | w_{<i}, h_t, h_v)$  is the probability of generating a token at the \textit{i}-th position given the previously generated tokens, input text, and image tokens. 
The visual tokens are incorporated with textual tokens and then fed to the language model. 

\subsection{Evolver: Chain-of-Evolution Prompting}
\label{sec:detail}

To improve LMMs' understanding of the online hateful memes which are evolving in nature, we design a novel \underline{C}hain-\underline{o}f-\underline{E}volution (CoE) prompt which has three components: (a) an evolutionary pair mining module to identify most relevant candidate memes; (b) an evolution information extractor to extract key information from the candidate memes; and (c) a contextual relevance amplifier for more effective hatefulness detection.
       
\myparagraph{Evolutionary Pair Mining}
It is unusual to find memes that have an evolutionary relationship with existing memes. Motivated by \citet{QHPBZZ23}, the evolution of a meme is defined as {\it new memes that emerge by fusing other memes or cultural ideas}. Therefore, the evolution of memes and old memes share similar textual and visual semantic regularities. We leverage this property to identify these old memes given the evolved meme. To this end, we can leverage the evolution of hateful memes to enhance LMMs' ability to understand hateful memes. In the implementation, we first generate the textual and visual embeddings from an external meme pool and target memes with CLIP. Specifically, the external meme pool should have two characteristics (1) do not overlap with the test set. (2) have enough evolutionary information. For simplicity, we use the training set as the carefully curated meme pools rather than any other dataset, where the memes follow the same definition of hatefulness/harm/misogyny. Then, we fuse the textual and visual embeddings with a fixed ratio. Ideally, meme evolution has temporal ability, which means that the memes change with time. However, it is difficult to accurately locate the upstream and downstream of meme evolution with existing technology. Therefore, we use multiple evolutionary memes to find the common characteristics of this evolution. Finally, given a target image-text pair, we retrieve the top-\textit{K} similar memes using cosine similarity:


\begin{gather}
    \text{memes} = \{A_i | \text{cos}(\mathbf{A}, \mathbf{B})_i \in \text{Top}_K(\text{cos}(\mathbf{A}, \mathbf{B})) \}
\end{gather} where $\mathbf{A} \in \mathcal{R}^{n \times d}$ is embedding of $n$ candidate memes and $\mathbf{B} \in \mathcal{R}^d$ is the  $d$-dimensional vector of the target meme. $\text{cos} (\cdot)$ return a similarity vector and $\text{Top}_K(\cdot)$ returns $K$ highest values given the input vector. For each evolution meme, we pair \textit{K} memes that the evolution meme is derived from. 

\noindent \myparagraph{Evolution Information Extractor}
To extract the information we are interested in (\eg \textit{hateful component}), we summarize paired memes with the help of a large multimodal model. The whole process can be expressed as follows:
\begin{equation}
    \text{Info} = \text{LMM}([\text{memes}_{top_K}, X_{extract}])
\end{equation}
where Info stands for our evolutionary information and LMM indicates the large multimodal model. $\text{memes}_{Top_K}$ are the $Top_K$ memes retrieved in the previous step, and $X_{extract}$ is the instruction to guide the LMM to extract information. We present the detailed instruction of $X_{extract}$ as shown in Table~\ref{tab:instruction}. More definitions of hateful memes from different datases are shown in Appendix~\ref{sec:prompt-R}.
\begin{table}[ht!]\centering
\begin{minipage}{0.99\columnwidth}\vspace{0mm}    \centering
\begin{tcolorbox} 
    \centering
   
     \hspace{-4mm}
      \scriptsize
    \begin{tabular}{p{0.99\columnwidth}}

Extract the common harmful feature of these image caption pairs based on the following hatefulness rules: 
\\
\\
\VarSty{\textbf{Any attacks on characteristics, including ethnicity, race, nationality, immigration status, religion, caste, sex, gender identity, sexual orientation, and disability or disease should be considered hateful. We define attack as violent or dehumanizing (comparing people to non-human things, e.g. animals) speech, statements of inferiority, and calls for exclusion or segregation. Mocking hate crime is considered hateful.}}
\\
\\
Input: [image 0 : <image0>, caption 0 : {texts[0]}, image 1 : <image1>, caption 1 : {texts[1]}, image 2 : <image2>, caption 2 : {texts[2]}, , image 3 : <image3>, caption 3 : {texts[3]}, image 4 : <image4>, caption 4 : {texts[4]}']
\\
\\           
Output: [Here is your response]

    \end{tabular}
\end{tcolorbox}
\vspace{-2mm}
\caption{The example of the instruction for extracting the evolutionary information.}
\label{tab:instruction}
\end{minipage}
\end{table}

\noindent \myparagraph{Contextual Relevance Amplifier}
To enhance the in-context hatefulness information, we add a contextual relevance amplifier to the LMM during evolution information extraction and final prediction. The contextual relevance amplifier can help increase the search for hateful components. In our case, the contextual relevance amplifier is the definition of a hateful meme given by the dataset we use. Finally, we combine the extracted information and contextual relevance amplifier as the in-context enhancement and feed them to the model: 
\begin{equation}
    \hat y = \text{LMM}([\text{memes}_{T}, X_{D}, \text{Info}, \text{Amp}])
\end{equation}
where Info stands for the information extracted previously, $\hat y$ is the final prediction, $\text{memes}_{T}$ is the memes we want to detect, and $X_{D}$ is the instruction to ask the large multimodal model to detect hateful memes. Amp refers to contextual relevance amplifier. The example of the amplifier is the same as the \VarSty{\textbf{blue}} part in Table~\ref{tab:instruction}.

\noindent \myparagraph{The Principles of Prompts} Our prompt strategy is simple yet effective, based on two key principles: (1) include the hateful meme definition and (2) limit the prompt to 30 words, addressing LMMs’ challenges with long-text comprehension. We directly applied definitions from the FHM, MAMI, and HarM datasets, summarized to meet this length requirement with the whole prompt using GPT-4.  To this end, we do not use any specialized prompt design, highlighting the robustness of our method.


\vspace{-0.4em}
\begin{table*}[t]
\small
  \setlength\tabcolsep{0.85mm}
\vskip 0.1in
  \centering
  \begin{tabular}{l|c|cc|cc|cc}
    \toprule
     \multirow{2}{*}{\textbf{Methods}}  & \multirow{2}{*}{\textbf{Model Size}} &  \multicolumn{2}{c|}{\textbf{Dataset: FHM}} &  \multicolumn{2}{c|}{\textbf{Dataset: MAMI}} & \multicolumn{2}{c}{\textbf{Dataset: HarM}} \\
    &  &  AUC $\uparrow$ & ACC $\uparrow$ & AUC $\uparrow$ & ACC $\uparrow$  & AUC $\uparrow$ & ACC $\uparrow$ \\
    \midrule
    \rowcolor{mygray} \multicolumn{8}{c}{\textit{Typical Models (full-Supervised)}} \\
    CLIP BERT~\cite{pramanick-etal-2021-momenta-multimodal} & $\textless$1B & 67.0  & 58.3 & 77.7 & 68.4  & 82.6 &  80.8  \\
    Text BERT~\cite{kiela2020hateful} & $\textless$1B & 66.1  & 57.1 & 74.5 & 67.4  & 81.4 & 78.7   \\
    Image-Region~\cite{kiela2020hateful} & $\textless$1B & 56.7  & 52.3 & 70.2 & 64.2  & 74.5 & 73.1   \\
    \midrule
    \rowcolor{mygray} \multicolumn{8}{c}{\textit{API-based LMM  (Zero-shot)}} \\
    Gemini-Pro-V~\cite{team2023gemini} & - & 66.0  & 65.7 & 74.5 & 74.5  & 71.3 & 76.2  \\
    GPT-4V~\cite{2023GPT4VisionSC} & - & 70.5 & 70.3 & - & - & -  & - \\
    \midrule
    \rowcolor{mygray} \multicolumn{8}{c}{\textit{Open-source LMM (Zero-shot)}} \\
    Openflamingo~\cite{awadalla2023openflamingo} & 7B & 57.0  & 56.4  & 56.8 & 56.8 &  51.7 & 55.8  \\
    LLaVA-1.5~\cite{liu2023improved} & 13B &  61.8  & 61.4  & 57.4 & 57.4  & 55.0 &  54.5 \\
    MMICL~\cite{zhao2023mmicl} & 11B & 59.9 & 60.4 & 67.3 & 67.3  & 52.1 &  63.8 \\
    MiniGPT-v2~\cite{zhu2023minigpt} & 7B & 58.8  & 59.1  & 62.3 & 62.3 &  57.1 & 60.3   \\
    BLIP-2~\cite{li2023blip} & 11B & 56.4 & 55.8  & 59.4 & 59.4 & 56.8 & 60.6  \\
    InstructBLIP~\cite{dai2023instructblip} & 13B & 59.6 & 60.1   & 64.1 & 64.1 &  55.7 & 60.1     \\
    \midrule
    \rowcolor{myblue}  Evolver (Ours) & 11B & \textbf{63.5} & \textbf{63.6} & \textbf{68.6} & \textbf{68.6}  & \textbf{67.7}  &  \textbf{65.5}\\
    \rowcolor{myblue}  Evolver$^\dag$  (Ours) & 13B & 62.3 & 62.5 & 59.9 &  59.9 & 59.3  &  57.3\\
    \bottomrule
  \end{tabular}
  \caption{\textbf{Comparison among different LMMs on zero-shot hateful meme detection benchmarks.}  Evolver and Evolver$^\dag$ are MMICL and LLaVA 1.5 with CoE prompting.  The API-based models have ethical considerations, we can not directly apply our CoE prompt above them. Best results in the open-source setting are highlighted in \textbf{bold}.}
\label{tab:main_res}
\vskip -0.1in
\end{table*}       


\makeatletter
  \newcommand\figcaption{\def\@captype{figure}\caption}
  \newcommand\tabcaption{\def\@captype{table}\caption}
\makeatother

\begin{figure*}[t]
\begin{minipage}[b]{0.37\textwidth}
\centering
\tabfootnotesize
\centering
\setlength{\tabcolsep}{8.5pt}
\begin{tabular}{cccc}
\toprule[1.25pt]
 \multirow{2}{*}{\textbf{EPM}} &  \multirow{2}{*}{\textbf{EIE}} &  \multirow{2}{*}{\textbf{CRA}} & \textbf{Datatset: FHM}  \\  
&  & & ACC$\uparrow$ \\ 
\midrule
& &  & 55.0  \\ \midrule
 \tiny{\Checkmark} & &  &  57.3 {\tiny \textbf{\textcolor{green}{(+2.3)}}} \\\
 & \tiny{\Checkmark} &  &  61.7 {\tiny \textbf{\textcolor{green}{(+6.7)}}} \\
 & \tiny{\Checkmark} & \tiny{\Checkmark}  &  60.6 {\tiny \textbf{\textcolor{green}{(+5.6)}}}  \\
\tiny{\Checkmark}  & \tiny{\Checkmark} & & 63.0 {\tiny \textbf{\textcolor{green}{(+8.0)}}}  \\
\rowcolor{myblue} \tiny{\Checkmark} & \tiny{\Checkmark} & \tiny{\Checkmark}  & \textbf{63.5} {\tiny \textbf{\textcolor{green}{(+8.5)}}} \\
\bottomrule[1.25pt]
\end{tabular}
\vspace{-.8em}
\tabcaption{\textbf{Ablation study of the three components} of our method on FHM dataset.}
\label{ablation}
  \end{minipage}
  \hfill
  \begin{minipage}[b]{0.58\textwidth}
    \centering
    \includegraphics[width=0.85\textwidth]{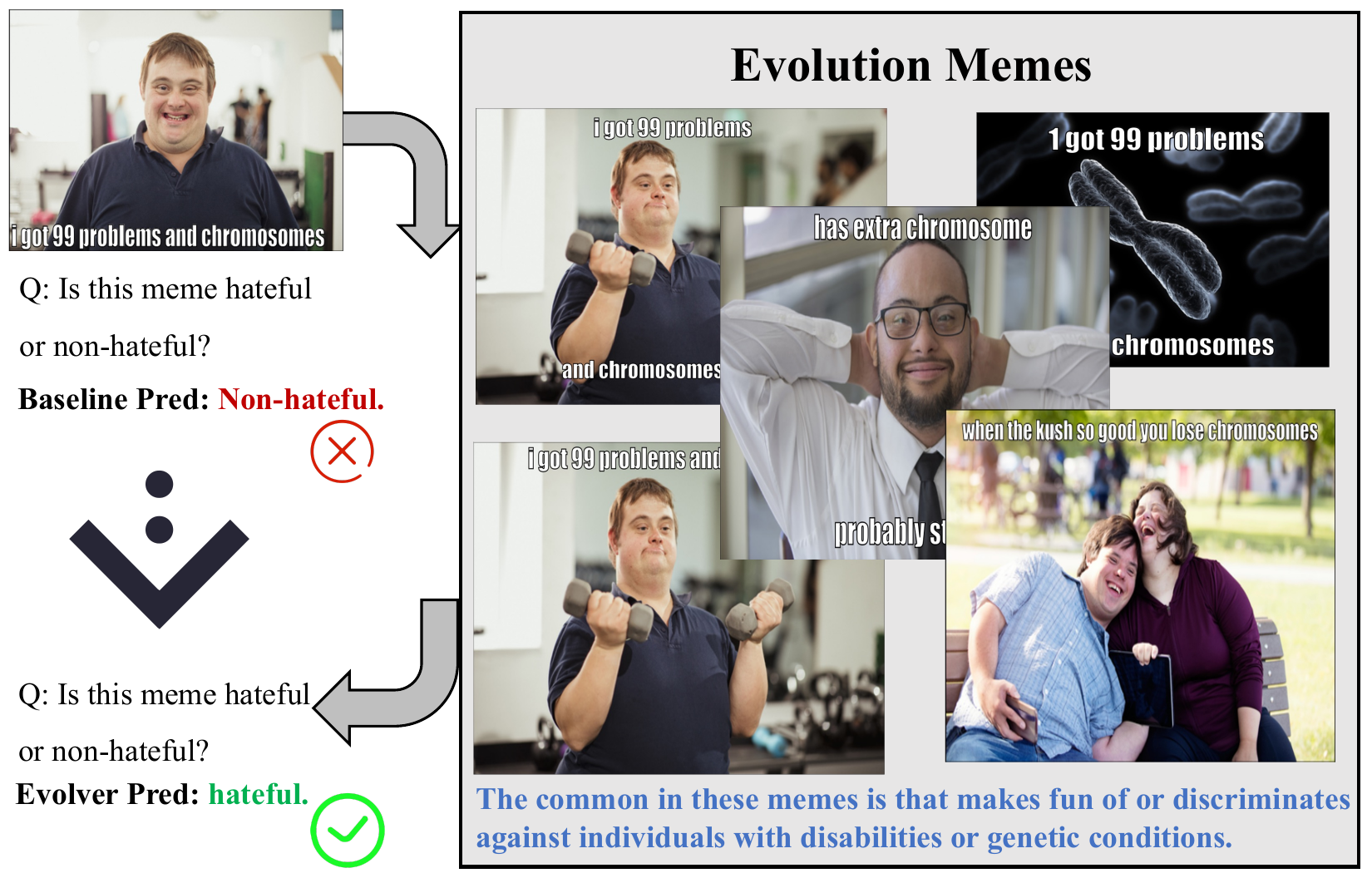}
    \vspace{-.8em}
    \caption{\textbf{Example results of the Evolver (Ours) and the baseline model (MMICL).} For more examples refer to the Appendix.}
\label{fig:visualization}
 \end{minipage}
\vspace{-0.3em}
\end{figure*}


\section{Experiment}

\subsection{Experimental Setup}
\myparagraph{Datasets} We evaluate \textit{Evolver} on three widely-used public datasets, namely, Facebook Hateful Meme dataset (FHM)~\cite{kiela2020hateful}, Harmful Meme dataset (HarM)~\cite{pramanick-etal-2021-momenta-multimodal}, and Multimedia Automatic Misogyny Identification (MAMI)~\cite{fersini-etal-2022-semeval}.

\noindent \myparagraph{Implementation}
We implement our \textbf{Evolver} based on the MMICL and LLaVA-1.5. For MMICL, we set the minimum length for the generation to 50, the maximum length for the generation to 80 during the Evolution Info Extractor, and the {\tt temperature} to 0.2. During the final prediction, we set the minimum length of generation to 1, and the maximum length for generation to 50. For LLaVA-1.5, we set {\tt temperature} to 0.2 and maximum generated tokens to 1024 for both stages. The embedding size of textual and visual embeddings is $N$ $\times$ 768, and $N$ is the number of memes. We fuse textual and visual embeddings with a fixed ratio of 4:1 by element-wise add in practice.

\noindent \myparagraph{Baselines}
The detailed description of the baseline models including API-based (\eg GPT-4V) and open-source LMMs (\eg LLaVA) can be found in Appendix~\ref{sec:baseline}. 

\noindent \myparagraph{Metrics} 
We adopt ACC (accuracy) and AUC (area under the ROC curve) as evaluation metrics.

\noindent \myparagraph{LMM Backbones}
We implement our \textit{Evolver} based on MMICL~\cite{zhao2023mmicl} and LLaVA-1.5~\cite{liu2023improved}.  
Please refer to the Appendix~\ref{sec:exp_detail} for more details.

\subsection{Main Result}
Table~\ref{tab:main_res} highlights the effectiveness of our Chain-of-Evolution Prompting strategy across three datasets. We compare zero-shot results among closed-source LMMs, open-source LMMs, and open-source LMMs with the CoE Prompting strategy. First, we observe that with more evolutionary context, LMMs exhibit a greater ability to recognize hateful content. Notably, both MMICL and LLaVA-1.5 with CoE Prompting outperform their zero-shot baselines, with LLaVA-1.5 achieving a 1.1\% improvement and MMICL a 3.2\% improvement in accuracy on the FHM dataset. Additionally, within open-source LMMs, model size does not necessarily lead to a better understanding of hateful memes. For instance, on MAMI, MiniGPT-v2 (7B) outperforms LLaVA-1.5 (13B). Furthermore, despite some closed-source LMMs being unavailable for evaluation on certain datasets due to model safety updates, these closed-source models still outperform open-source LMMs in zero-shot results, suggesting a significant gap for open-source models to close. The typical existing training-based models are fully supervised settings, mainly based on CLIP~\cite{radford2021learning} and BERT~\cite{devlin2018bert}, which can not provide interpretability. However, our setting is a zero-shot setting,  mainly including the LMM-based methods, which can explicitly reflect the evolution of memes and training-free. The superior performance of Gemini-Pro-V and GPT-4V is due to their larger parameter count and extensive training data. Furthermore, without security restrictions, our method could be seamlessly integrated into these models.

\subsection{Ablation Study}
In Table~\ref{ablation}, evolutionary pair mining (EPM) boosts the baseline with an improvement of up to 2.3\%. Moreover, the evolutionary information extractor (EIE) and contextual relevance amplifier (CRA) significantly improve the generalization ability. Our full model achieves the best performance and outperforms the baseline by 8.5\% in accuracy. This demonstrates that three parts of \textit{Evolver} are beneficial by integrating with the evolution information.



\vspace{-0.6em}
\subsection{Qualitative Analysis}
We visualize the paired evolution memes retrieved by the evolutionary pair mining modules in Figure~\ref{fig:visualization}, showing how the evolution information influences the final prediction.  The origin model (MMICL) without considering the meme's evolution, predicts non-hateful because it is not easy to detect the boy's disability or genetic conditions. After obtaining the evolution information like ``discriminates against individuals with disabilities or genetic conditions.", 
our {\it Evolver} rectify the prediction, which supports the rationale of our method. 

\subsection{Impact of Number of Evolutionary Memes}
\begin{figure}[ht]
\centering
    \includegraphics[width=1.0\linewidth]{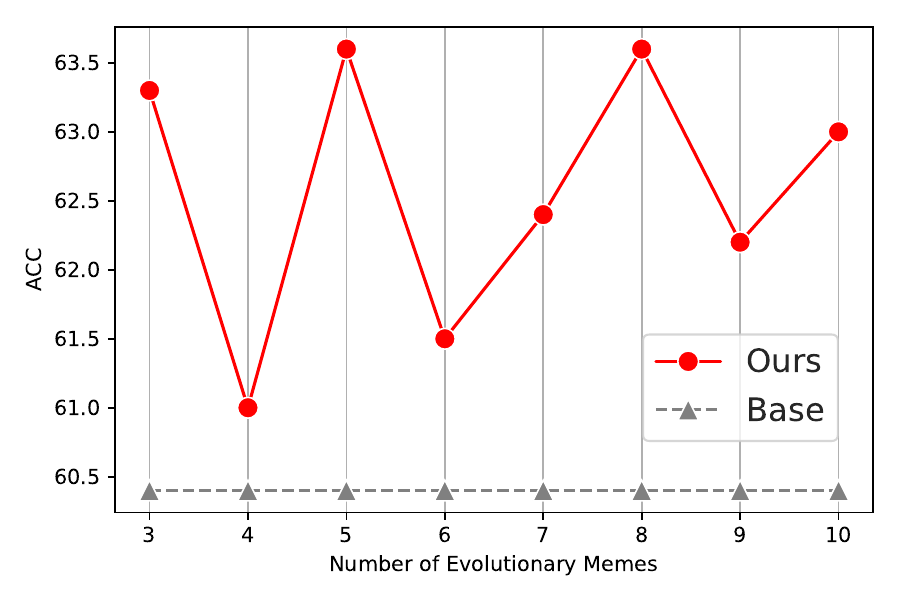}   
\vskip -0.1in
\caption{Effect of the number of evolutionary memes.}
\label{fig:abs_num}
\vskip -0.15in
\end{figure}

As shown in Figure~\ref{fig:abs_num}, we show the effect of the number $K$ of the evolutionary memes. We find that our method significantly improves the baseline under all evolutionary meme number settings. Specifically, we set the $K=5$ to achieve the best performance in practice.


\section{Conclusion}
We present \textit{Evolver} to seamlessly boost LMMs for hateful meme detection via Chain-of-Evolution prompting. By integrating evolution of memes, our method can adapt to unseen memes. Experimental results demonstrate the effectiveness of \textit{Evolver}.
\section*{Impact Statements}
\noindent \myparagraph{Ethics Statement.} 
While our objective is to mitigate the spread and impact of hate speech online, we recognize the potential for misuse or unintended consequences of this technology. 
Our framework is designed exclusively for detecting hateful memes, with its use strictly limited to academic or approved research environments. It is not intended for content generation. Insights from this research can support public awareness campaigns, encouraging informed digital citizenship and empowering users to actively participate in recognizing and reporting harmful content.
We are committed to ensuring that our technology is used ethically and responsibly.

\noindent \myparagraph{Reproducibility Statement.} We have clarified inference details including hyper-parameters, and the chain-of-evolution pipeline in Section~\ref{sec:detail} and Appendix. In addition, all the datasets used in this paper are open-source and can be accessed online.

\noindent \myparagraph{Limitation.} While \textit{Evolver} demonstrates significant advances in detecting hateful memes by leveraging the evolution of memes and large multimodal models (LMMs), it has some limitations. First, the effectiveness of our approach relies heavily on the quality and diversity of the curated meme pool used for seeking evolutionary memes. Furthermore, biases inherent in these datasets could potentially affect the model's ability to generalize across different cultural contexts and meme evolution patterns not represented in the related data. 

\newpage

\bibliography{custom}
\appendix
\twocolumn[{
\begin{center}
\large 
\end{center}
\bigskip
}]

\section{Baseline Details} 
\label{sec:baseline}

To ensure a fair comparison, we create the LMM-based zero-shot hateful speech detection benchmark.
(a) For API-based LMMs, we select GPT-4V and Gemini-Pro-V, which are among the most popular models.
(b) For open-source LMMs, our comparison extends to include six widely used models: Openflamingo, LLaVA-1.5, MMICL, MiniGPTv2, BLIP-2, and InstructBLIP. 

\noindent \myparagraph{GPT-4V} \cite{2023GPT4VisionSC} first creates a GPT-based LMM on the massive volume of datasets on the internet. They trained the language decoder, then aligned it with the vision encoder. 

\noindent \myparagraph{Gemini-Pro-V} \cite{team2023gemini} curates a large multimodal model with various image-text data on the web and then trains the model with image-text data from scratch.

\noindent \myparagraph{Openflamingo} \cite{awadalla2023openflamingo} builds a LMM using CLIP as vision encoder and MPT/RedPajama as a language decoder. The model is trained on open-source image-text datasets.

\noindent \myparagraph{LLaVA 1.5} \cite{liu2023improved} improves the performance of the model they propose previously \cite{liu2023visual} where CLIP is used as vision encoder and Vicuna as language decoder, with MLP projection to bridge the gap between vision encoder and language decoder. 


\noindent \myparagraph{MiniGPTv2 } \cite{zhu2023minigpt} connects the vision and language space with Q-former and an MLP projection. The model is pre-trained with a large collection of aligned image-text pairs and then instruction fine-tuned on curated high-quality image-text
pairs.

\noindent \myparagraph{BLIP-2 } \cite{li2023blip} propose a model architecture using only Q-former to connect the vision encoder and language decoder to perform a series of multi-modal tasks. The pre-training of model has two stages. 


\noindent \myparagraph{InstructBLIP } \cite{dai2023instructblip} further performs instruction fine-tuning based on BLIP-2 various converted multi-model datasets. 

\noindent \myparagraph{MMICL} \cite{zhao2023mmicl} propose a framework 
using Q-former and an MLP layer to connect a vision encoder and language decoder. 

\section{Data Details} 
\label{sec:exp_detail}




For more clarity, we show the detailed hateful meme detection dataset analysis in Table~\ref{tab:sum_dataset}. Here, we give a detailed description of each dataset. 

\begin{table}[h]
\centering
\tabfootnotesize
\setlength{\tabcolsep}{10.0pt}
\begin{tabular}{lcc}
\hline
\textbf{Dataset} & \textbf{\# of Train Samples} & \textbf{\# of Test Samples} \\
\hline
FHM & 8,500 & 1,000 \\
MAMI & 9,948 & 1,000 \\
HarM& 3,013 & 354 \\
\hline
\end{tabular}
\caption{Summary of hateful meme detection datasets.}
\label{tab:sum_dataset}
\end{table}

\noindent \textbf{FHM}~\cite{kiela2020hateful} curate a dataset of various hateful memes collected on the web to help build models to detect hateful memes. As shown in Table \ref{tab:sum_dataset}, it has 8,500 training examples and 1,000 test examples. The definition of a hateful meme is: 
\begin{table}[h!]\centering
\begin{minipage}{0.99\columnwidth}\vspace{0mm}    \centering
\begin{tcolorbox} 
    \centering
     \hspace{-4mm}
      \scriptsize
    \begin{tabular}{p{0.99\columnwidth}}
"A direct or indirect attack on people based on characteristics, including ethnicity, race, nationality, immigration status, religion, caste, sex, gender identity, sexual orientation, and disability or disease. We define attack as violent or dehumanizing (comparing people to non-human things, e.g. animals) speech, statements of inferiority, and calls for exclusion or segregation. Mocking hate crime is also considered hate speech."
    \end{tabular}
\end{tcolorbox}
\vspace{-2mm}
\caption{The definition of hatefulness in the FHM.}
\label{tab:def_1}
\end{minipage}
\end{table}

\noindent \textbf{MAMI}~\cite{fersini-etal-2022-semeval} propose a dataset of Misogyny memes, which contains 9,948 training examples as well as 1,000 test examples. The definition of Misogyny given by \textbf{MAMI} is: 
\begin{table}[h!]\centering
\begin{minipage}{0.99\columnwidth}\vspace{0mm}    \centering
\begin{tcolorbox} 
    \centering
     \hspace{-4mm}
      \scriptsize
    \begin{tabular}{p{0.99\columnwidth}}
"meme is misogynous if it conceptually describes an offensive, sexist, or hateful scene (weak or strong, implicitly or explicitly) having as target a woman or a group of women. Misogyny can be expressed in the form of shaming, stereotype, objectification, and/or violence."
    \end{tabular}
\end{tcolorbox}
\vspace{-2mm}
\caption{The definition of hatefulness in the MAMI.}
\label{tab:def_2}
\end{minipage}
\end{table}

\noindent \textbf{HarM}~\cite{pramanick-etal-2021-momenta-multimodal} builds a dataset of harmful meme related to Covid. As shown in Table \ref{tab:sum_dataset}, this dataset has 3,013 training samples and 354 test examples. Additionally, \textbf{HarM} gives the definition of a harmful meme as follows: 

\begin{table}[h!]\centering
\begin{minipage}{0.99\columnwidth}\vspace{0mm}    \centering
\begin{tcolorbox} 
    \centering
     \hspace{-4mm}
      \scriptsize
    \begin{tabular}{p{0.99\columnwidth}}
"Multi-modal unit consisting of an image and an embedded text that has the potential to cause harm to an individual, an organization, a community, or society"
    \end{tabular}
\end{tcolorbox}
\vspace{-2mm}
\caption{The definition of hatefulness in the HarM.}
\label{tab:def_3}
\end{minipage}
\end{table}
\vspace{-2mm}
\section{Computing Resource Requirements}
Our Evolver is seamlessly integrated into the LMMs in the inference state, requiring minimal computational resources. For the main results, we conduct experiments on one GTX-3090-24G or one A100-40G. For the ablation study, we measure the comparison on one A100-40G.


\section{Prompt Engineering}
\label{sec:prompt-R}

In this section, we show the detailed prompt of Evolution Information Extractor (EIE) for \textbf{MAMI} and \textbf{HarM} datasets in Table~\ref{tab:EIE_1} and Table~\ref{tab:EIE_2}.

\begin{table}[h!]\centering
\begin{minipage}{0.99\columnwidth}\vspace{0mm}    \centering
\begin{tcolorbox} 
    \centering
      \scriptsize
    \begin{tabular}{p{0.99\columnwidth}}

Extract the common misogynous feature of these image caption pairs based on the following misogynous rules: 

\VarSty{\textbf{meme is misogynous if it conceptually describes an offensive,
sexist, or hateful scene (weak or strong, implicitly or explicitly)
having as target a woman or a group of women. Misogyny can
be expressed in the form of shaming, stereotype, objectification,
and/or violence.}}
\end{tabular}
\end{tcolorbox}
\vspace{-2mm}
\caption{The prompt design for EIE in the MAMI.}
\label{tab:EIE_1}
\end{minipage}
\end{table}

\vspace{-4mm}
\begin{table}[h!]\centering
\begin{minipage}{0.99\columnwidth}\vspace{0mm}    \centering
\begin{tcolorbox} 
    \centering
     \hspace{-4mm}
      \scriptsize
\begin{tabular}{p{0.99\columnwidth}}

Extract the common harmful feature of these image caption pairs based on the following misogynous rules: 

\VarSty{\textbf{meme is harmful if Multi-modal unit consisting of an image and an embedded text that
has the potential to cause harm to an individual, an organization, a
community, or society.}}

\end{tabular}
\end{tcolorbox}
\vspace{-2mm}
\caption{The prompt design for EIE in the HarM.}
\label{tab:EIE_2}
\end{minipage}
\end{table}

Furthermore, we show the detailed prompt of our Evolver in the final prediction for \textbf{MAMI} and \textbf{HarM} datasets in Table~\ref{tab:Final_1} and Table~\ref{tab:Final_2}.

\begin{table}[h!]\centering
\begin{minipage}{0.99\columnwidth}\vspace{0mm}    \centering
\begin{tcolorbox} 
    \centering
     \hspace{-4mm}
      \scriptsize
    \begin{tabular}{p{0.99\columnwidth}}

Determine if an image <image0> with its caption: $ocr_{text}$ is misogynous or not misogynous. Below is the misogynous analysis of 5 evolutional image text pairs, where our target image text pair is derived from those evolution image text pairs.
Use the misogynous of evolution to help you make decision.  

Evolution: <Info>

Requirement: \\
1. The output should be a word of either 'misogynous' or 'not misogynous'.
2. meme is misogynous if it conceptually describes an offensive,
sexist, or hateful scene (weak or strong, implicitly or explicitly)
having as target a woman or a group of women. Misogyny can
be expressed in the form of shaming, stereotype, objectification,
and/or violence.

[Here is your expert response]

\end{tabular}
\end{tcolorbox}
\vspace{-2mm}
\caption{Prompt design for final prediction of MAMI.}
\label{tab:Final_1}
\end{minipage}
\end{table}
\vspace{-4mm}

\begin{table}[ht!]\centering
\begin{minipage}{0.99\columnwidth}\vspace{0mm}    \centering
\begin{tcolorbox} 
    \centering
   
     \hspace{-4mm}
      \scriptsize
    \begin{tabular}{p{0.99\columnwidth}}

Determine if an image <image0> with its caption: $ocr_{text}$ is harmful or not harmful. Below is the harmfulness analysis of 5 evolutional image text pairs, where our target image text pair is derived from those evolution image text pairs.
Use the harmfulness of evolution to help you make decision.  

Evolution: <Info>

Requirement: \\
1. The output should be a word of either 'harmful' or 'not harmful'.
2. meme is misogynous if Multi-modal unit consisting of an image and an embedded text that has the potential to cause harm to an individual, an organization, a community, or society.

[Here is your expert response]

\end{tabular}
\end{tcolorbox}
\vspace{-2mm}
\caption{Prompt design for final prediction of HarM.}
\label{tab:Final_2}
\end{minipage}
\vspace{-7mm}
\end{table}

\section{Related Work}
\myparagraph{Hateful Meme Detection.} Most works of hateful meme detection are based on pretrained vision-language models and fine-tune them on the hateful meme detection data. For example, \citet{kiela2020hateful} introduced the Hateful Memes Challenge and proposed a multimodal model that combines visual and textual features using a transformer-based architecture. Hate-CLIPper proposed a multimodal contrastive learning approach to improve the representation learning of memes by aligning the visual and textual modalities. These works demonstrate the effectiveness of using pretrained vision-language models. Differently, we discuss the potential of LMMs for hateful meme detection and introduce a knowledge-enhance LMM.

\noindent \myparagraph{Large Multimodal Models.} 
With the release of the GPT-4V~\cite{2023GPT4VisionSC}, many researchers focus on the development and application of LMMs in processing and understanding multimodal data. Discuss groundbreaking models like GPT-4V and LLAVA-1.5~\cite{liu2023improved}, highlighting their capabilities in interpreting complex datasets that include a combination of text and images. Address the challenges these models face, such as sensitivity to noise and typographical errors, and the ongoing efforts to improve their accuracy and robustness in real-world applications. 
In this work, we adopt the LMMs to hateful meme detection, where set an LMM-based zero-shot hateful meme detection benchmark. Furthermore, we show that our \textit{Evolver} further boosts the LMM's performance. 

\noindent \myparagraph{Prompt-based Learning.}
Prompt-based learning has evolved significantly from its inception with large language models (LLMs) to its current application in various tasks. The concept of prompt engineering emerged with the release of GPT-3, which demonstrated that LLMs could be guided to generate desired outputs through carefully crafted prompts. Inheriting the merit from LLM, prompt-based learning in large multimodal models (LLMs) has also gained extensive research attention. In our work, we present a novel chain-of-evolution prompting strategy to seamlessly boost the existing LMM for hateful meme detection. 

\begin{figure*}[ht!]
    \centering
    \includegraphics[width=1.0\textwidth]{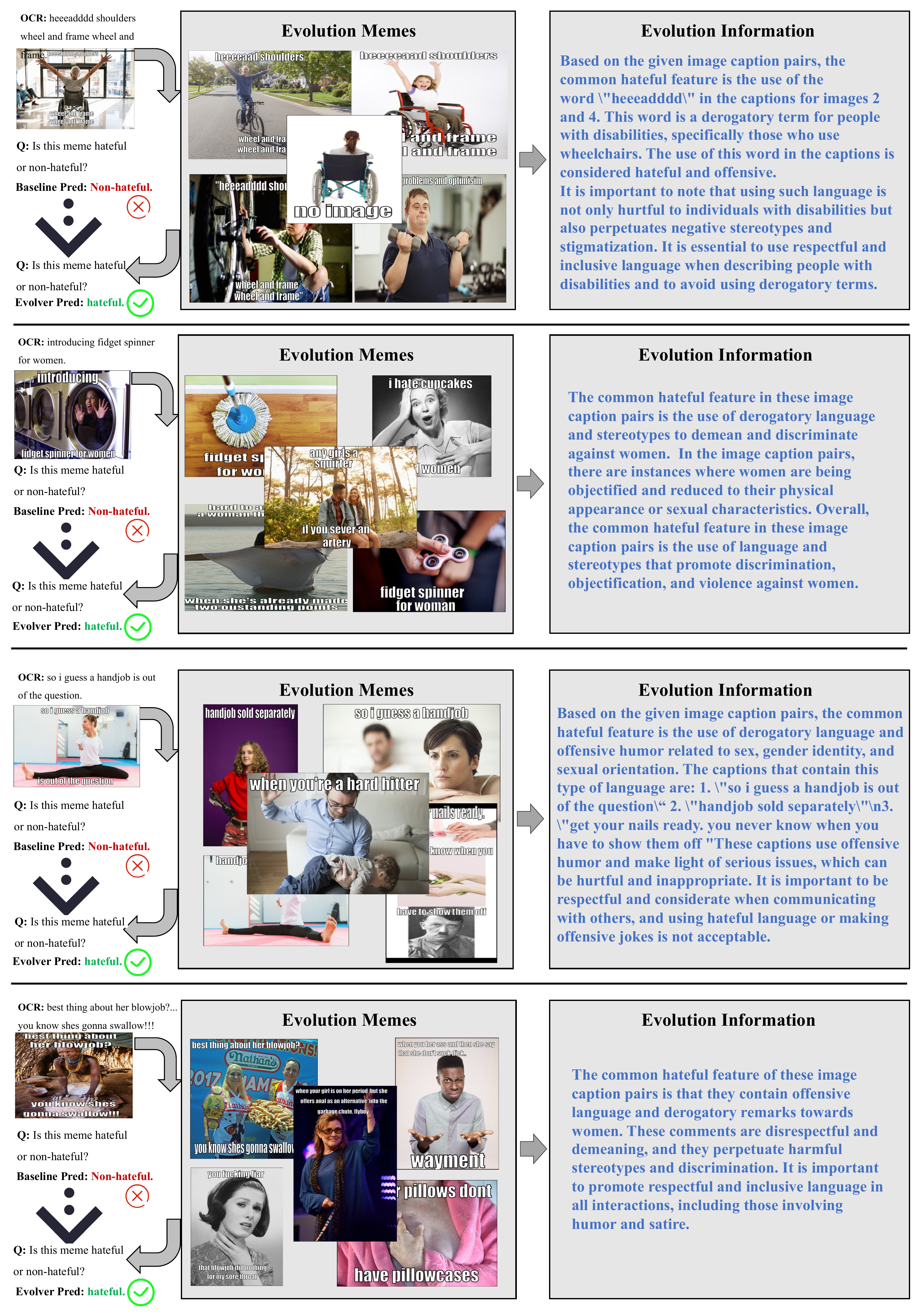}
    \caption{Example results of the Evolver (Ours) and the baseline model (MMICL).}
    \label{fig:addition1}
    \vspace{-10pt}
\end{figure*}

\noindent \myparagraph{More Cases Analysis.} 
To show the effectiveness of our method and align a better understanding of hatefulness in FHM dataset, we show more examples generated by \textit{Evolver} and baseline in Fig.~\ref{fig:addition1}.


\end{document}